\documentclass{article}
\usepackage{spconf,amsmath,amssymb,graphicx,booktabs,array,url}
\usepackage[T1]{fontenc}
\usepackage{microtype,flushend,needspace}
\title{\fontsize{14}{16}\selectfont MSR-ABC: Complementary Feature Fusion for Clear Cell Renal Cell Carcinoma Classification}
\name{Yuan Liang$^{1,3}$, Sourav Bhattacharjee$^{2}$, and Abraham Campbell$^{3}$}
\address{\fontsize{10}{11.5}\selectfont
$^{1}$Research Ireland Centre for Research Training in Machine Learning\\
\fontsize{10}{11.5}\selectfont $^{2}$School of Veterinary Medicine, University College Dublin, Dublin, Ireland\\
\fontsize{10}{11.5}\selectfont $^{3}$School of Computer Science, University College Dublin, Dublin, Ireland\\
\fontsize{10}{11.5}\selectfont yuan.liang@ucdconnect.ie}
\begin{document}
\ninept
\maketitle
\begin{abstract}
Clear cell renal cell carcinoma (ccRCC) classification from computed tomography (CT) may benefit from representations that capture different aspects of tumour phenotype. However, simply concatenating reconstruction-derived features, image--text embeddings, and radiomic descriptors may not fully exploit their complementary information. We propose MSR-ABC, a global--local framework integrating MedVAE, MedSigLIP, and radiomics. Spatially matched local image descriptors interact through cross-representation attention, while radiomic auxiliary supervision encourages the pooled local representation to retain quantitative tumour attributes. A gated residual mechanism integrates local information with global summaries. Area-aware pooling and visual and soft-text prompting complete the framework. We evaluate discrimination using a patient-level split of $396$ KiTS23 cases and an external cohort of $111$ TCGA/TCIA cases, reporting mean AUROC over three training seeds with $95\%$ patient-bootstrap confidence intervals. MSR-ABC achieves mean AUROCs of $0.806$ internally and $0.879$ externally. Internally, it exceeds global MedVAE--radiomics and three-source concatenation by $3.35$ and $3.61$ AUROC percentage points, respectively. Removing any source lowers mean AUROC in both cohorts, with the largest decreases following radiomics removal. Individual removal of cross-representation attention, auxiliary supervision, or gated fusion also lowers internal mean AUROC. These findings are consistent with complementary contributions within the integrated architecture and support further investigation of structured representation fusion for non-invasive renal tumour characterisation.

\end{abstract}

\section{Introduction}
Clear cell renal cell carcinoma (ccRCC) is the predominant RCC subtype. Histological subtype contributes prognostic information alongside tumour stage and grade~\cite{eau}. CT depicts tumour morphology and enhancement, but imaging phenotypes overlap across histological categories. Quantitative subtype assessment before treatment could therefore complement clinical interpretation. A multicentre study has investigated multiphase CT radiomics for renal tumour subtype assessment~\cite{uhlig}.

Radiomics (R) encodes predefined intensity, shape, and texture descriptors. Learned features provide another route to image characterisation. MedVAE (M) compresses medical images through large-scale autoencoder pretraining~\cite{medvae}; MedSigLIP (S) maps images and text into a shared medical representation space~\cite{medgemma}. These objectives motivate combining reconstruction features, image--text representations, and explicit tumour measurements. This is complementarity between representations of the same CT examination, rather than between separate clinical imaging modalities.

Global concatenation does not explicitly link matched local M/S regions, supervise their phenotype content, or control their contribution to the case-level decision. MSR-ABC addresses these three aspects within one global--local model.

We propose \emph{MSR-ABC}: \textbf{A} interacts spatially matched M/S tokens, \textbf{B} provides radiomic auxiliary supervision, and \textbf{C} gates local information into global features. Area-aware pooling, visual prompt tuning (VPT), and soft-text prompting complete the implementation. Source-removal and architectural controls assess internal Test discrimination and its transfer to the external cohort.

\section{Method}
\begin{figure*}[t]
\centering
\includegraphics[width=\textwidth]{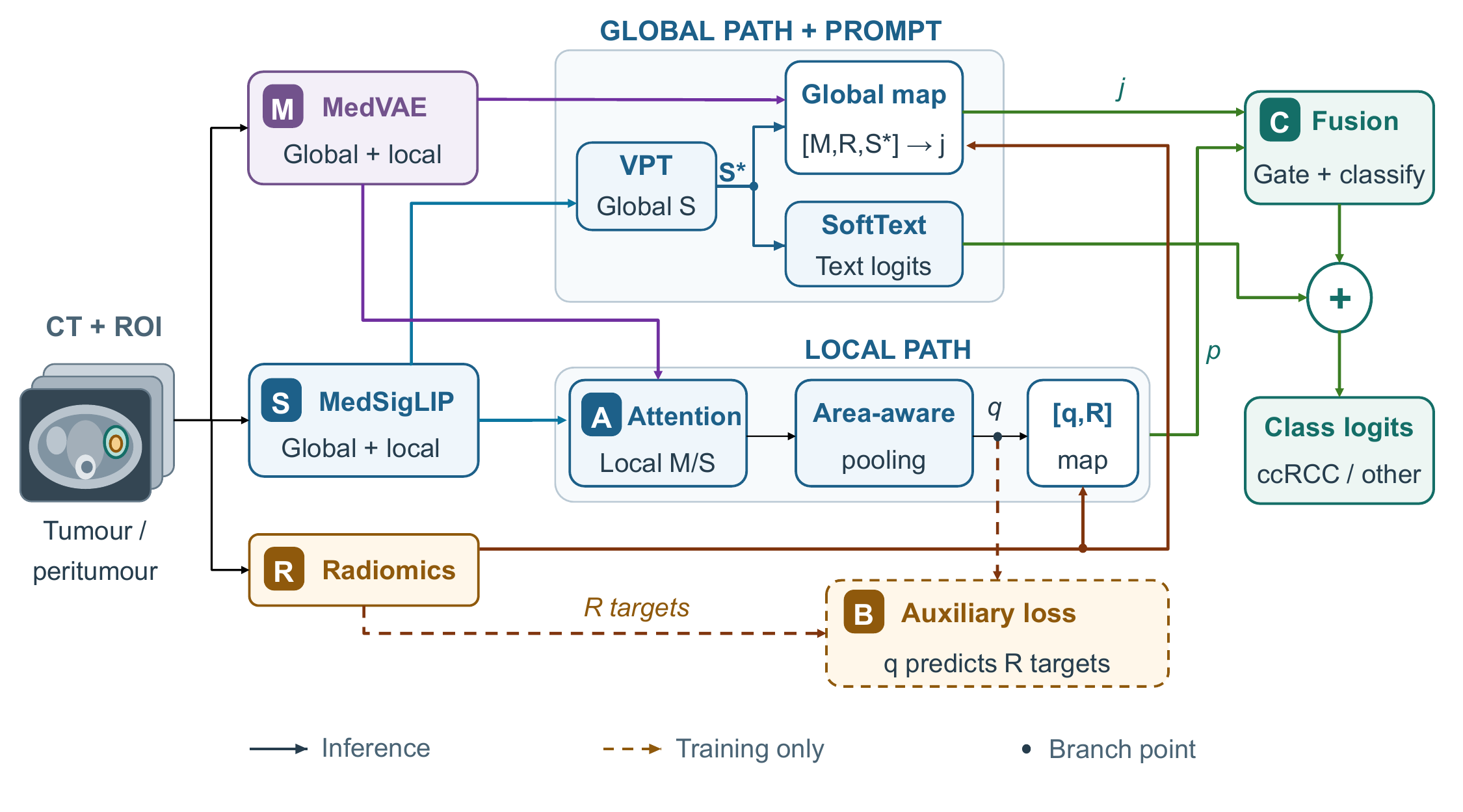}
\caption{MSR-ABC architecture. The CT illustration is schematic. $S^*$ denotes VPT-adapted global S, $j$ the global summary, $q$ the pooled local image summary, and $p$ the mapped $[q,R]$ representation. Brackets denote concatenation after preprocessing. The circled plus adds fusion and image--text logits. Solid and dashed arrows denote inference and training-only paths. Dots mark branches. Crossings without dots are unconnected. B backpropagation is omitted.}
\label{fig:architecture}
\end{figure*}
\subsection{Cohorts and representation construction}
The internal cohort includes 396 cases from the KiTS23 dataset~\cite{kits,kits_paper}, divided at the patient level into approximately 60\% training, 25\% validation, and 15\% Test. Each split contains approximately 70\% ccRCC and 30\% non-ccRCC cases. The external cohort comprises 111 TCGA/TCIA cases~\cite{kirc,kirp,kich}, with approximately 77\% ccRCC and 23\% non-ccRCC. Segmentation annotations were obtained from RCC-AID~\cite{rccaid,rccaid_paper} and the AIMI Annotations initiative~\cite{aimi}.

\textbf{Global versus local paths.} Fig.~\ref{fig:architecture} shows how MSR-ABC combines CT images and region-of-interest (ROI) masks in one model with two paths. The global path combines patient-level M/R/S summaries. The local path first combines spatially matched M/S descriptors and then pools them to a patient-level representation. R enters both paths as a case-level descriptor.

The fixed global M descriptor comprises 32 training-fitted principal components of frozen MedVAE tumour/peritumoural features and a region-validity indicator (33 values). R contains 73 descriptors obtained by label-free screening: inter-annotator ICC(3,1) $\geq0.75$, removal of the lowest 20\% by median absolute deviation (MAD), Pearson redundancy filtering at $|r|>0.95$, and average-linkage clustering of $1-|r|$ at distance 0.2, retaining the highest-MAD descriptor in each cluster. The global S embedding averages five tumour-rich axial slices, after per-slice normalisation, into a normalised 1,152-dimensional representation; fixed training-fitted scaling and PCA yield 32 coordinates. Concatenating M, R, and S gives 138 inputs.

For the local pathway, a tumour-centred $96^3$ context is sampled using physical geometry, with intensities clipped to $[-200,300]$ HU. Twelve depth bins contribute tumour-area-selected planes, each divided into a $2\times2$ grid. Tumour interior (T) and the 0--10 mm peritumoural region within the kidney (N) define up to 96 spatial slots per encoder. Fixed local M and S descriptors are each reduced to 16 dimensions and projected to 32-dimensional tokens, with spatial-coordinate, region, and encoder embeddings. Spatial slots without valid tissue receive zero pooling weight.

\subsection{ABC integration and area-aware pooling}
\textbf{A: local cross-representation interaction.} At each valid spatial slot, the matched 32-dimensional M and S tokens enter a four-head self-attention block with residual connections, layer normalisation, a 32--64--32 feed-forward network, and dropout probability 0.1. The two output tokens exchange information between encoders at that location before spatial aggregation. Attention does not mix different spatial slots.

\textbf{Area-aware pooling.} Gated attention pooling~\cite{mil} aggregates the updated tokens separately within T and N. Each learned attention score receives the logarithm of the token's regional fractional pixel mass, with coefficient one, before softmax normalisation. Thus, larger tissue coverage increases the weight for otherwise equal learned scores. Matched M/S tokens share the area term. Two pooled 32-vectors and two availability flags are mapped to the local image summary $q\in\mathbb{R}^{64}$. Invalid slots receive zero weight.

\textbf{B: radiomic auxiliary supervision.} An auxiliary multilayer perceptron reads $q$, before R is concatenated, and predicts 16 training-selected radiomic attributes. These R targets undergo fixed inverse-hyperbolic-sine transformation and normalisation. All selected training targets are available. A Huber loss with transition one and weight 0.1 supplements class-weighted cross-entropy. Its gradients update the local branch producing $q$, encouraging image-derived features to retain radiomic information. B receives R only as supervision, and its predictions do not enter C. This auxiliary head is used only during training.

\Needspace{7\baselineskip}\textbf{C: global--local gated residual fusion.} A linear/GELU map projects the 138 preprocessed $[M,R,S^*]$ inputs to the global representation $j\in\mathbb{R}^{64}$. A separate map combines pooled local information $q$ with the same case-level R to give $p\in\mathbb{R}^{64}$. Both are patient-level representations. Their integration is
\begin{equation}
z=j+\sigma\!\left(W_g[j;p]+b_g\right)\odot p,
\label{eq:gate}
\end{equation}
Here, $z$ is the fused representation, $W_g$ and $b_g$ are learned gate weights and bias, $\sigma$ is sigmoid, and $\odot$ denotes elementwise multiplication. Using both $j$ and $p$, the gate scales each component of $p$ between zero and one while retaining $j$. A 64--32--2 classifier maps $z$ to two fusion logits.

\subsection{Visual and soft-text adaptation}
\textbf{VPT: adapting the global image representation.} Following visual prompt tuning~\cite{vpt}, four learnable 1,152-dimensional tokens enter before the final (27th) frozen MedSigLIP vision block. They adapt global image features while pretrained weights remain fixed. Pooling retains only the original 1,024 image tokens, yielding the VPT-adapted 1,152-dimensional global embedding $S^*$. For global fusion, its change relative to the unprompted embedding is projected through fixed PCA/scaling and added to the original 32-dimensional S representation. SoftText uses $S^*$ before this projection. Local S remains unchanged.

\textbf{SoftText: adding class-text evidence.} Following learnable context prompting~\cite{coop}, four shared context vectors precede fixed ccRCC/non-ccRCC descriptions in a 64-token sequence, preserving the end-of-sequence position. The frozen text encoder produces two class embeddings. Their cosine similarities with $S^*$ are scaled by ten and mapped through a zero-initialised, bias-free $2\times2$ residual layer. These image--text scores, labelled Text logits in Fig.~\ref{fig:architecture}, are added to the fusion logits at the circled plus. The resulting Class logits are pre-softmax scores for ccRCC and other (non-ccRCC) cases.

\subsection{Training, controls, and statistical reporting}
Models use AdamW (learning rate $10^{-3}$), with checkpoints selected by validation AUROC.

We report mean individual-model AUROC over three seeds and percentile 95\% confidence intervals (CIs) from 50,000 class-stratified patient bootstrap resamples shared across methods.

Global-only baselines comprise an M-only MLP, MR/MSR concatenation, and radiomics logistic regression (R-LR). R-LR uses the same number of radiomic features (73), selected independently on the training set, with balanced class weights and validation-selected L2 regularisation ($C=0.01$).

Source ablations retrain without the corresponding global/local inputs. Removing S also disables VPT and SoftText, and removing R disables B.

Component controls replace A with a tokenwise MLP, disable B, or replace C with a concatenation projection. Area ablation removes only the mass term. The main comparisons and ablations are summarised in Tables~\ref{tab:main} and~\ref{tab:ablation}, respectively.

\section{Results \& Discussion}
\subsection{Internal discrimination and external behaviour}
Table~\ref{tab:main} summarises the principal models. MSR-ABC reaches mean Test AUROC 0.8064 (95\% CI 0.658--0.927), compared with 0.7729 for global MR and 0.7703 for global MSR concatenation. The two global baselines have close internal point estimates, with no mean gain from simply appending S. The higher full-model estimate favours the combined global--local integration and prompting pipeline. M-only and R-LR have lower internal and external point estimates. Their contrast suggests that reconstruction-derived appearance alone is insufficient internally and supports including explicit tumour measurements.

\begin{table}[!ht]
\centering
\caption{Mean AUROC [95\% patient-bootstrap CI] over three seeds. R-LR is a deterministic single-model reference. Prompt rows add SoftText, VPT, or both to ABC + area. Area denotes area-aware pooling.}
\label{tab:main}
\fontsize{9}{10.6}\selectfont
\setlength{\tabcolsep}{2.4pt}
\begin{tabular}{@{}lcc@{}}
\toprule
Model & Test ($n=60$) & External ($n=111$) \\
\midrule
Global M & 0.5922 [0.443, 0.737] & 0.8335 [0.763, 0.895] \\
R-LR & 0.7513 [0.589, 0.892] & 0.8597 [0.781, 0.927] \\
Global MR & 0.7729 [0.623, 0.898] & 0.8792 [0.811, 0.937] \\
Global MSR & 0.7703 [0.616, 0.902] & 0.8893 [0.822, 0.945] \\
ABC + area & 0.7897 [0.641, 0.914] & 0.8691 [0.795, 0.931] \\
+ SoftText & 0.7897 [0.641, 0.915] & 0.8701 [0.796, 0.932] \\
+ VPT & 0.8007 [0.652, 0.924] & 0.8805 [0.811, 0.939] \\
\midrule
MSR-ABC (full) & 0.8064 [0.658, 0.927] & 0.8790 [0.809, 0.938] \\
\bottomrule
\end{tabular}
\end{table}

\Needspace{12\baselineskip}
\subsection{Contributions of the M, S, and R sources}
Table~\ref{tab:ablation} provides source and component controls. Removing R produces the largest internal source-removal decrease, consistent with explicit tumour measurements contributing information not reliably recovered by the frozen image representations. However, this also removes B supervision, and uncertainty remains substantial.

\begin{table}[!ht]
\centering
\caption{Source and component ablations: three-seed mean AUROC [95\% patient-bootstrap CI]. CIs are for AUROC, not differences. Full denotes MSR-ABC, and w/o means without. Source removals include dependent functionality.}
\label{tab:ablation}
\fontsize{9}{10.6}\selectfont
\setlength{\tabcolsep}{2.4pt}
\begin{tabular}{@{}lcc@{}}
\toprule
Variant & Test ($n=60$) & External ($n=111$) \\
\midrule
Full & 0.8064 [0.658, 0.927] & 0.8790 [0.809, 0.938] \\
w/o M source & 0.7897 [0.645, 0.911] & 0.8374 [0.756, 0.909] \\
w/o S source & 0.7937 [0.646, 0.915] & 0.8564 [0.781, 0.921] \\
w/o R source & 0.7288 [0.568, 0.872] & 0.8142 [0.729, 0.890] \\
w/o A & 0.7972 [0.649, 0.921] & 0.8735 [0.802, 0.934] \\
w/o B & 0.7994 [0.650, 0.922] & 0.8784 [0.809, 0.937] \\
w/o C & 0.7941 [0.645, 0.918] & 0.8683 [0.794, 0.931] \\
w/o area prior & 0.7998 [0.651, 0.923] & 0.8793 [0.810, 0.938] \\
\bottomrule
\end{tabular}
\end{table}

Removing either M or S lowers mean AUROC in both cohorts. For S, internal mean AUROC falls from 0.8064 to 0.7937, while external mean AUROC falls from 0.8790 to 0.8564. The consistent direction favours retaining the S-dependent pathway. Together with the M and R removals, this pattern is consistent with complementary contributions from the three source pathways within the integrated model. The S-removal effect cannot be assigned specifically to the embedding, VPT, or text residual because all are removed together. Global MSR concatenation still attains a higher external point estimate than the full model. The source-removal findings therefore support complementarity within this architecture without establishing its superiority over every alternative fusion design.

\subsection{Roles and limits of the ABC components}
Removing A, B, or C lowers mean internal AUROC. A allows M and S to exchange information before spatial pooling, rather than processing each token independently. Its higher mean AUROC relative to the tokenwise MLP in both cohorts supports including cross-representation interaction in the local pathway. C provides adaptive integration of that pathway with global features, and its removal gives the largest internal decrease among the component controls.

B imposes an auxiliary learning objective on the image-derived local representation without adding an inference-time prediction branch. The small AUROC decreases after removing B in both cohorts are consistent with a modest regularisation benefit. These results favour retaining both local interaction and auxiliary supervision within the full model. The descriptive differences do not establish the specific representation mechanisms. Area weighting adds a region-coverage bias, with a small internal mean advantage and a slightly lower external point estimate.

VPT and SoftText adapt the frozen representation path. The full model exceeds the no-prompt and single-prompt variants internally, including a mean AUROC advantage of 0.0057 over VPT alone. External AUROC remains slightly lower than VPT alone. The point estimates favour combined prompting internally, but do not show an advantage across both cohorts or establish visual--text synergy.

\section{Conclusion}
MSR-ABC combines reconstruction-derived, image–text, and radiomic representations through local interaction, auxiliary supervision, and gated global–local fusion. By linking spatially matched image features with patient-level summaries, the framework provides a structured approach to integrating different aspects of tumour phenotype. The three-seed point estimates favour the full model over simple concatenation and individual attention, supervision, or fusion ablations internally. Removing MedVAE, MedSigLIP, or radiomics lowers mean AUROC in both cohorts, consistent with complementary contributions from the complete source pathways within this architecture. External comparisons nevertheless indicate that gains depend on the cohort and reference model. Clinically, this CT-based characterisation could support non-invasive preoperative assessment and provide complementary information as an adjunct to radiological and pathological evaluation. Fresh-cohort validation, calibration, and evaluation of decision consequences are required to establish clinical utility before changing patient management.

\section{Compliance with Ethical Standards}
This retrospective study used publicly available KiTS23 and TCGA/TCIA data, including RCC-AID and AIMI annotations. No new participants were recruited or identifiable data collected. Ethical approval was not required for this secondary analysis of public data, as confirmed by the authors.

\section{Acknowledgments}
 No funding was received for this study. The authors have no competing interests to declare that are relevant to the content of this article.

\bibliographystyle{IEEEbib}
\bibliography{references}

@article{uhlig, author={Annemarie Uhlig and others}, title={Radiomics and machine learning for renal tumor subtype assessment using multiphase computed tomography in a multicenter setting}, journal={European Radiology}, volume={34}, number={10}, pages={6254--6263}, year={2024}, doi={10.1007/s00330-024-10731-6}}

@inproceedings{medvae, author={Maya Varma and others}, title={{MedVAE}: Efficient Automated Interpretation of Medical Images with Large-Scale Generalizable Autoencoders}, booktitle={Proc. MIDL}, series={PMLR}, volume={301}, pages={1597--1626}, year={2026}}

@article{medgemma, author={Andrew Sellergren and others}, title={{MedGemma} Technical Report}, journal={arXiv:2507.05201}, year={2025}}

@misc{kits, author={{KiTS23 Organizers}}, title={The 2023 Kidney and Kidney Tumor Segmentation Challenge}, howpublished={\url{https://kits-challenge.org/kits23/}}, year={2023}, note={Accessed Nov. 3, 2026}}

@misc{kirc, author={O. Akin and others}, title={The Cancer Genome Atlas Kidney Renal Clear Cell Carcinoma Collection ({TCGA-KIRC}), ver. 3}, howpublished={The Cancer Imaging Archive}, year={2016}, note={doi:10.7937/K9/TCIA.2016.V6PBVTDR}}

@misc{kirp, author={M. Linehan and others}, title={{TCGA-KIRP}: Kidney Renal Papillary Cell Carcinoma Collection, ver. 4}, howpublished={The Cancer Imaging Archive}, year={2016}, note={doi:10.7937/K9/TCIA.2016.ACWOGBEF}}

@misc{kich, author={M. W. Linehan and R. Gautam and C. A. Sadow and S. Levine}, title={The Cancer Genome Atlas Kidney Chromophobe Collection ({TCGA-KICH}), ver. 3}, howpublished={The Cancer Imaging Archive}, year={2016}, note={doi:10.7937/K9/TCIA.2016.YU3RBCZN}}

@misc{rccaid, author={Sarah de Boer and others}, title={{RCC-AID}: Renal Cell Carcinoma AI Dataset for Medical Imaging Research}, howpublished={Zenodo}, year={2026}, note={doi:10.5281/zenodo.19630297}}

@misc{aimi, author={Jeff Van Oss and Gowtham Krishnan Murugesan and Diana McCrumb and Rahul Soni}, title={Image segmentations produced by the {AIMI} Annotations initiative}, howpublished={Zenodo}, year={2023}, note={Version 1.4. doi:10.5281/zenodo.8428938}}

@inproceedings{mil, author={Maximilian Ilse and Jakub M. Tomczak and Max Welling}, title={Attention-based Deep Multiple Instance Learning}, booktitle={Proc. ICML}, series={PMLR}, volume={80}, pages={2127--2136}, year={2018}}

@inproceedings{vpt, author={Menglin Jia and others}, title={Visual Prompt Tuning}, booktitle={Computer Vision -- ECCV 2022}, series={LNCS}, volume={13693}, pages={709--727}, year={2022}, doi={10.1007/978-3-031-19827-4_41}, note={doi:10.1007/978-3-031-19827-4\_41}}

@article{coop, author={Kaiyang Zhou and Jingkang Yang and Chen Change Loy and Ziwei Liu}, title={Learning to Prompt for Vision-Language Models}, journal={International Journal of Computer Vision}, volume={130}, pages={2337--2348}, year={2022}, doi={10.1007/s11263-022-01653-1}}

@misc{eau, author={{European Association of Urology}}, title={{EAU} Guidelines on Renal Cell Carcinoma}, howpublished={\url{https://uroweb.org/guidelines/renal-cell-carcinoma}}, year={2026}, note={Chapters 3 and 6. Accessed Nov. 3, 2026}}

@article{kits_paper, author={Nicholas Heller and others}, title={The {KiTS21} Challenge: Automatic segmentation of kidneys, renal tumors, and renal cysts in corticomedullary-phase {CT}}, journal={arXiv:2307.01984}, year={2023}, doi={10.48550/arXiv.2307.01984}}

@article{rccaid_paper, author={Sarah de Boer and others}, title={{RCC-AID}: Renal Cell Carcinoma AI Dataset for Medical Imaging Research}, journal={Machine Learning for Biomedical Imaging}, volume={2026}, pages={772--782}, year={2026}, doi={10.59275/j.melba.2026-2dg2}, note={doi:10.59275/j.melba.2026-2dg2}}
\end{document}